\documentclass[journal]{IEEEtran}
\usepackage{url}
\usepackage{array}
\usepackage{amsfonts}
\usepackage{amsmath}
\usepackage{amssymb}
\usepackage{booktabs}
\usepackage{cite}
\usepackage{color}
\usepackage{multirow}
\usepackage{MnSymbol}
\usepackage{microtype}
\usepackage{graphicx}
\usepackage{subfigure}
\usepackage{booktabs} 
\usepackage{parskip}

\usepackage{atbegshi}
\AtBeginDocument{\AtBeginShipoutNext{\AtBeginShipoutDiscard}}

\begin{document}
%
\title{XFlow: Cross-modal Deep Neural Networks\\ for Audiovisual Classification}

%
%
%
\author{C\u{a}t\u{a}lina~Cangea, Petar~{Veli{\v c}kovi{\'c}}, and~Pietro~Li\`o}
\thanks{C. Cangea, P. Veli{\v c}kovi{\'c} and P. Li\`o are with the Department
of Computer Science and Technology, University of Cambridge, Cambridge CB03DF, United Kingdom, e-mails: \{ccc53, pv273, pl219\}@cst.cam.ac.uk.}

\maketitle
\setcounter{page}{1}
\begin{abstract}
In recent years, there have been numerous developments towards solving multimodal tasks, aiming to learn a stronger representation than through a single modality. Certain aspects of the data can be particularly useful in this case---for example, correlations in the space or time domain across modalities---but should be wisely exploited in order to benefit from their full predictive potential. We propose two deep learning architectures with multimodal cross-connections that allow for dataflow between several feature extractors (XFlow). Our models derive more interpretable features and achieve better performances than models which do not exchange representations, usefully exploiting correlations between audio and visual data, which have a different dimensionality and are nontrivially exchangeable. Our work improves on existing multimodal deep learning algorithms in two essential ways: (1) it presents a novel method for performing cross-modality (before features are learned from individual modalities) and (2) extends the previously proposed cross-connections which only transfer information between streams that process compatible data. Illustrating some of the representations learned by the connections, we analyse their contribution to the increase in discrimination ability and reveal their compatibility with a lip-reading network intermediate representation. We provide the research community with Digits, a new dataset consisting of three data types extracted from videos of people saying the digits 0--9. Results show that both cross-modal architectures outperform their baselines (by up to 11.5\%) when evaluated on the AVletters, CUAVE and Digits datasets, achieving state-of-the-art results.
\end{abstract}


%
\IEEEpeerreviewmaketitle

\section{Introduction}
%
%
%
%

An interesting extension of unimodal learning consists of deep models which ``fuse'' several modalities (for example, sound, image or text) and thereby learn a shared representation, outperforming previous architectures on discriminative tasks. However, the cross-modality in existing models using restricted Boltzmann machines~\cite{ngiam2011multimodal}, deep Boltzmann machines~\cite{srivastava2012multimodal} and similarity-based loss functions in deep convolutional networks~\cite{deepalignedrepresentations} only occurs after the unimodal features are learned. This prevents the unimodal feature extractors from exploiting any information contained within the other modalities. The work presented in this paper has focused on direct information exchange between the unimodal feature extractors, while deriving more interpretable features, therefore making it possible to directly exploit the correlations between modalities. This information exchange may occur between data of varying dimensionality (for example, 1D/2D for audiovisual data) and thus poses a highly nontrivial problem.

Cross-connections were previously introduced by {Veli{\v c}kovi{\'c}} et al.~\cite{2016arXiv161000163V} to obtain better performance on sparse datasets (when limited numbers of samples are available) through directly exploiting the heterogeneity of the available features. The cross-connections achieve this by allowing information to be exchanged between hidden layers of neural networks that are each processing a subset of the input data. Each constituent network will consequently learn the target function from exactly one of these subsets. Partitioning the data helps each superlayer achieve better predictive performance by reducing the dimensionality of the input. However, each of the networks is processing data that is compatible with the other networks.

To the best of our knowledge, there are no multimodal learning algorithms capable of transforming and exchanging features between learning streams in a modular and flexible manner. As discussed in the previous paragraph, cross-connections have been effective in improving the classification performance on sparse datasets by passing feature maps between constituent networks. We hypothesise that predictive tasks involving multimodal data can benefit from a generalised cross-connection approach, primarily in domains where the different modalities are aligned and highly correlated---for example, in the audiovisual data domain. Our proposed method is motivated by the plentiful existence of correlations in audio and visual streams from speech recordings, which can lead to a stronger joint representation of the corresponding signals. These alignments should be exploited before the feature extraction phase has ended, so devising a generalised method of feature passing between learning streams seems like a natural approach.

In this study, we present cross-connections that are capable of feature exchange between 1- and 2-dimensional signals and can be, in principle, generalised to data types of any dimensionality. We validate their effectiveness in significantly improving model performance on audiovisual classification tasks, showing that cross-modal feature exchanges are beneficial for the learning streams of a multimodal architecture. Our contributions are as follows:

\begin{figure*}[t!]
\begin{center}
\includegraphics[height=2in]{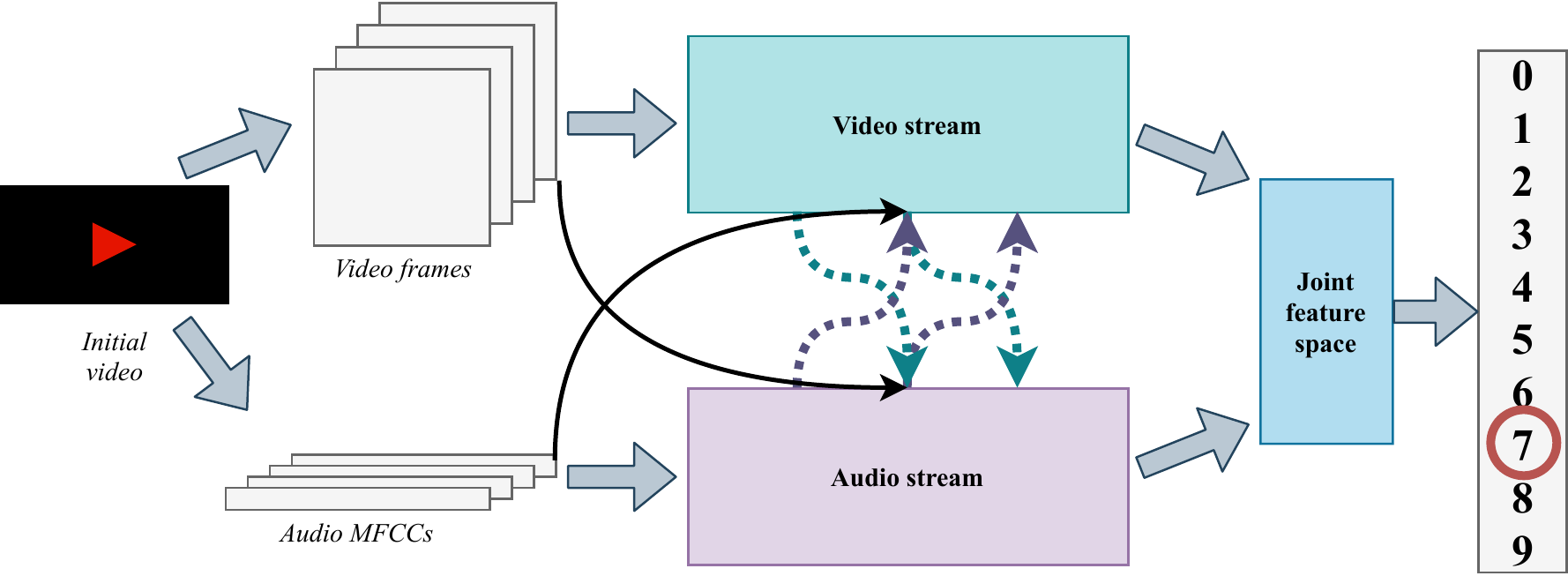}
\caption{High-level graphical description of our proposed multimodal classification system. In order to determine which letter/digit a person is saying, we provide the visual and audio signals as input to two processing streams. The latter extract features from the modalities while reinforcing their intermediate representations. This is achieved by appending information received from the other stream via two kinds of connections (internal and residual cross-modal connections, represented by dashed and straight lines, respectively). Finally, the resulting features are joined and the class decision is made.}
\label{fig:xflow_overall}
\end{center}
\vskip -0.1in
\end{figure*}

\begin{itemize}
	\item Generalisable (1D$\rightsquigarrow$2D and 2D$\rightsquigarrow$1D) cross-modal connections that exploit the correlations between audio and image and two models for multimodal speech classification that incorporate these connections. Our models obtain a significant edge over their corresponding baselines (identical models without cross-modal connections), illustrating that a better representation can be learned when fusion between the modalities takes place during the feature extraction process. To illustrate the vast improvement, we present results which show that these models achieve state-of-the-art results on the \emph{AVletters} and \emph{CUAVE} benchmark tasks. We illustrate the high-level structure and data flow within our architectures in Figure~\ref{fig:xflow_overall}.

	\item In addition to the improved methods for addressing cross-modality tasks, in order to address the issues present within existing datasets, we have also constructed \emph{Digits}---a novel, open dataset which is of superior quality to other existing benchmark multimodal audiovisual datasets\footnote{The dataset will be publicly released upon publication.}. With 750 examples belonging to 10 classes, \emph{Digits} contains three different data types (video frames, audio coefficients and spectrograms) which can provide researchers with various possibilities of validating their future approaches to multimodal machine learning.

	\item Finally, the existence of cross-connections has allowed for a more direct way to analyse the correlations present within different input modalities. As the last part of our contribution, we directly analyse these representations, deriving useful conclusions about their mutual constructiveness for the classification task at hand. This is a step towards addressing the interpretability issues encountered by deep learning models.
\end{itemize}

\section{XFlow Models}

\subsection{CNN $\times$ MLP}

\begin{figure}[h]
\begin{center}
\includegraphics[clip, trim=1cm 4cm 3cm 0cm, width=0.4\textwidth]{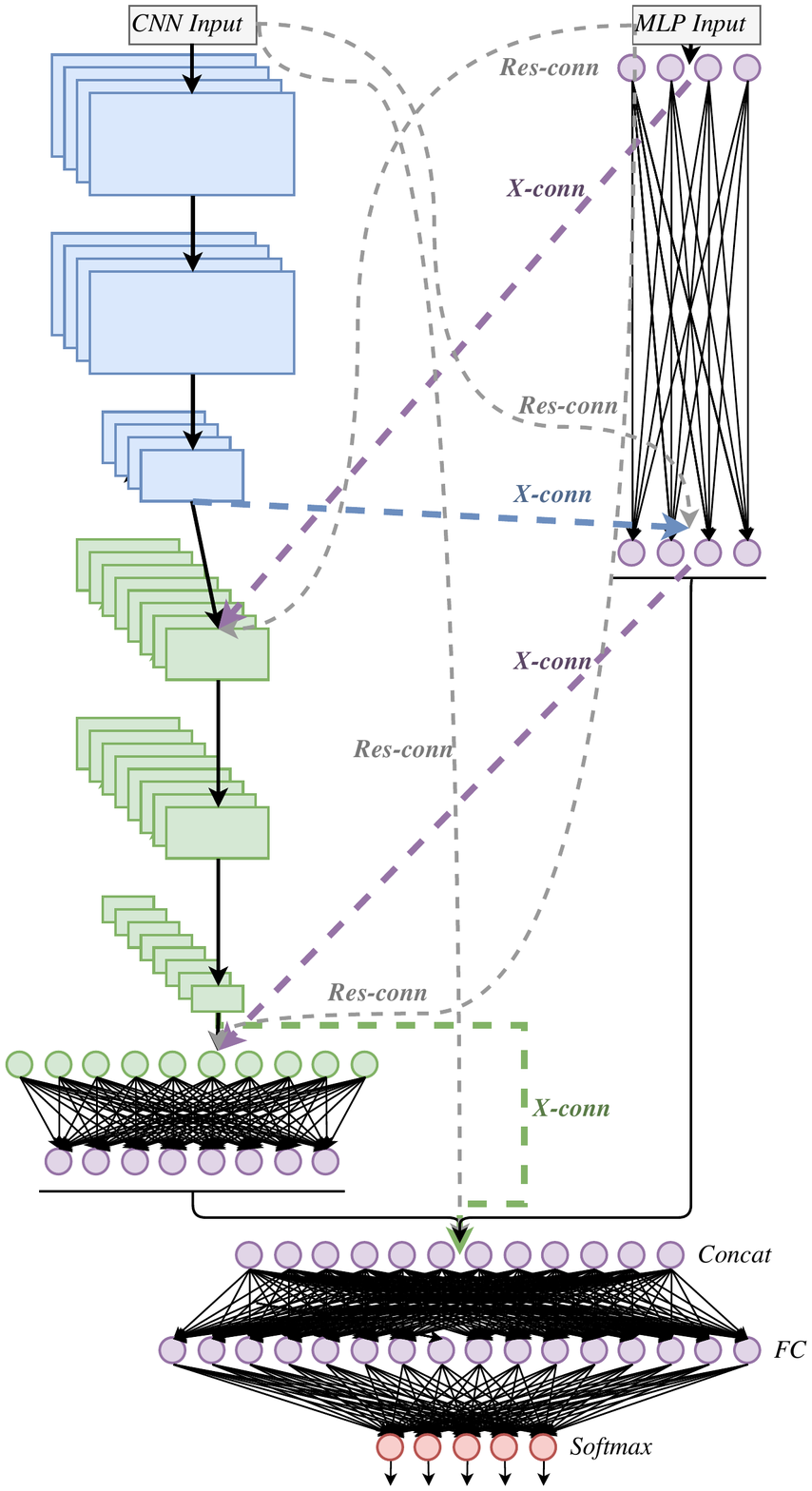}
\caption{CNN $\times$ MLP model with cross- and residual connections, denoted by X-conn and Res-conn, respectively. The former are shown in thick dashes, while the latter are illustrated using thin dashes.}
\label{fig:cnn_mlp_x}
\end{center}
\vskip -0.1in
\end{figure}

Illustrated in Figure~\ref{fig:cnn_mlp_x}, the first multimodal architecture takes as input a tuple \texttt{(x\_img, x\_mfcc)} and outputs a probability distribution over the possible classes that this example belongs to. The first element represents a 2D visual modality (the video frames for a person saying a letter) and is processed by a convolutional neural network, whereas the second one consists of 1D audio data corresponding to the same frames, in the form of mel-frequency cepstral coefficients (MFCCs),  and is fed into a multi-layer perceptron. Once features are extracted separately, they are concatenated and passed through another MLP for classification.

This architecture can only process fixed-size inputs, so we had to perform sliding window averaging over video frames and corresponding MFCC sets for all examples in the dataset. Since the length of a video in an example can vary considerably from person to person, some examples had to undergo averaging over a large window size. This results in loss of information about the transitions between consecutive frames, and we anticipated that this would hurt the performance of the model.

In order to exploit correlations between audio and visual information, the model leverages cross- and residual connections. Both of these represent cross-modal sequences of neural network layers---they transform data representations from 1D to 2D and vice versa, passing the resulting features to the other processing stream, in an end-to-end learnable fashion. This ensures that the network will learn effective transformations via cross-modal connections, which can benefit the target streams and thereby improve the task performance. Furthermore, we can choose to cross-connect multiple times and between points of varying depth, essentially allowing the network to choose representations which are most beneficial to the task performance. The input to cross-connections (which are illustrated with thick dashes in Figure~\ref{fig:cnn_mlp_x}) is an intermediate representation from the source learning stream, whereas cross-modal residual connections (thin dashes) operate directly on the raw input to the stream. The outputs of connections are merged with intermediate representations at the corresponding point in the target stream.

\subsubsection{Cross-connections}

\begin{figure}[h]
\begin{center}
\includegraphics[clip, trim=1cm 4.5cm 2cm 0cm, width=0.37\textwidth]{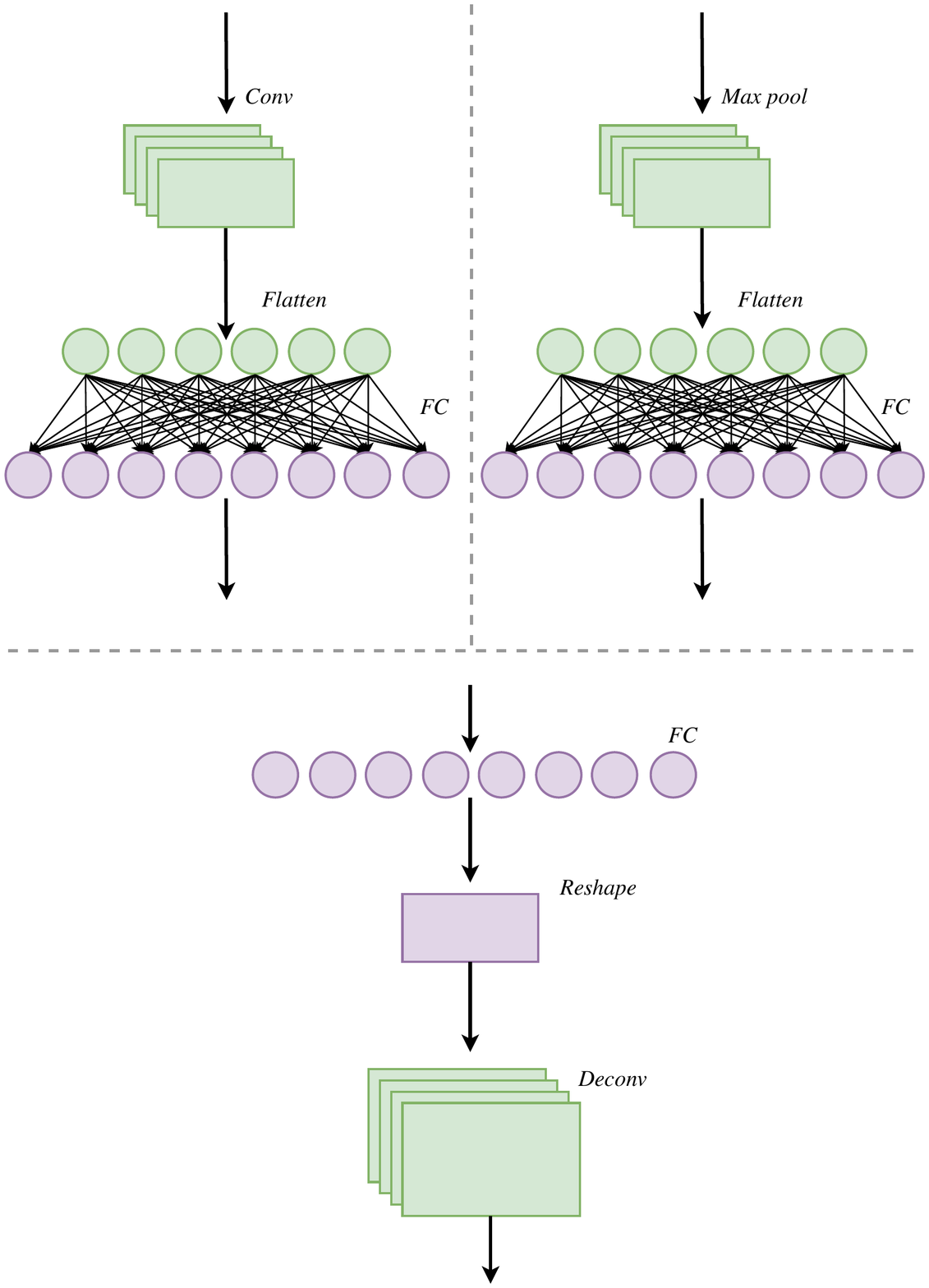}
\caption{Detailed view of the connections. \textit{(Upper left:)} 2D$\rightsquigarrow$1D cross-connection. \textit{(Upper right:)} 2D$\rightsquigarrow$1D residual connection. \textit{(Bottom:)} 1D$\rightsquigarrow$2D cross-/residual connection.}
\label{fig:conns_cnn_mlp_x}
\end{center}
\vskip -0.1in
\end{figure}

The primary role of \textit{cross-connections} (shown in Figure~\ref{fig:conns_cnn_mlp_x}) is to perform information exchange while the features from individual modalities are being learned (before the concatenation operation). A fundamental incompatibility exists between 1D and 2D data---there is no trivially interpretable way of transferring the feature maps resulting from a \{conv$\times 2$, max-pool\} block to the fully-connected layers processing the audio data and vice versa. We therefore had to design more complex types of cross-connections that would enable the data to be exchanged in a sensible manner and allow useful interpretations of these transfers.

The 2D$\rightsquigarrow$1D cross-connections (depicted by blue and green thick dashes in Figure~\ref{fig:cnn_mlp_x}) have the following structure: the output of a \{conv$\times 2$, max-pool\} block in the CNN is passed through a convolutional layer. The result is then flattened and processed by a fully-connected layer. Finally, we concatenate the output of the latter with the output of the corresponding fully-connected layer in the MLP (for the first cross-connection) or directly with the outputs of the CNN and MLP (for the second one).

The 1D$\rightsquigarrow$2D cross-connections (shown in purple thick dashes in Figure~\ref{fig:cnn_mlp_x}) perform the inverse operation: the output of a fully-connected layer is passed through another layer of the same type, such that the number of features matches the dimensionality required for the deconvolution operation. We apply the latter to the reshaped data and concatenate the result with the output of the corresponding \{conv$\times 2$, max-pool\} block. To summarise, we provide the output of the fully-connected layer to a reshape operation followed by a deconvolution---this sequence of operations forces the network to learn during the training process how to ``translate'' the 1D signal characteristics, essentially enabling the structure to be present in the resulting 2D representation.

\subsubsection{Cross-modal residual connections}

Residual learning~\cite{he2016deep} has the purpose of making the internal layers in a neural network represent the data more accurately. Our cross-connection design allows for straightforwardly including residual cross-modal connections that allow the raw input of one modality to directly interact with another modality's intermediate representations. This effectively has the potential to correct for any unwanted effects that one stream's intermediate transformations might have caused. Figure~\ref{fig:conns_cnn_mlp_x} also illustrates residual connections, constructed in a similar manner to cross-connections.

Mathematically, a 1D$\rightsquigarrow$2D cross-connection with residuals will add $\textit{Reshape}(\mathbf{W}_{\text{res}}\mathbf{x}_{\text{in}} + \mathbf{b}_{\text{res}}) * \textbf{K}_{\text{res}}$ to the 2D stream at depth $d$ and concatenate $\textit{Reshape}(\mathbf{W}_{\text{xcon}}\mathbf{h}_d + \mathbf{b}_{\text{xcon}}) * \textbf{K}_{\text{xcon}}$, where $\mathbf{W}_*$ and $\mathbf{K}_*$ are learnable weights, $\mathbf{b}_*$ are learnable biases, $\mathbf{x}_{\text{in}}$ are inputs and $\mathbf{h}_d$ are intermediate layer outputs.

\subsection{\{CNN $\times$ MLP\}--LSTM}\label{lstmarch}

The second architecture processes the same kind of data as the CNN $\times$ MLP model, namely tuples of the form \texttt{(x\_img, x\_mfcc)}. However, the fundamental difference lies in the fact that each video frame/MFCCs pair is being provided separately as input to the pre-concatenation streams. This brings forward the crucial advantage of not having to average the data across more frames, keeping the temporal structure intact and maintaining a richer source of features from both modalities.

\begin{figure}[h]
\begin{center}
\includegraphics[clip, trim=2cm 18cm 5cm 4cm, width=0.47\textwidth]{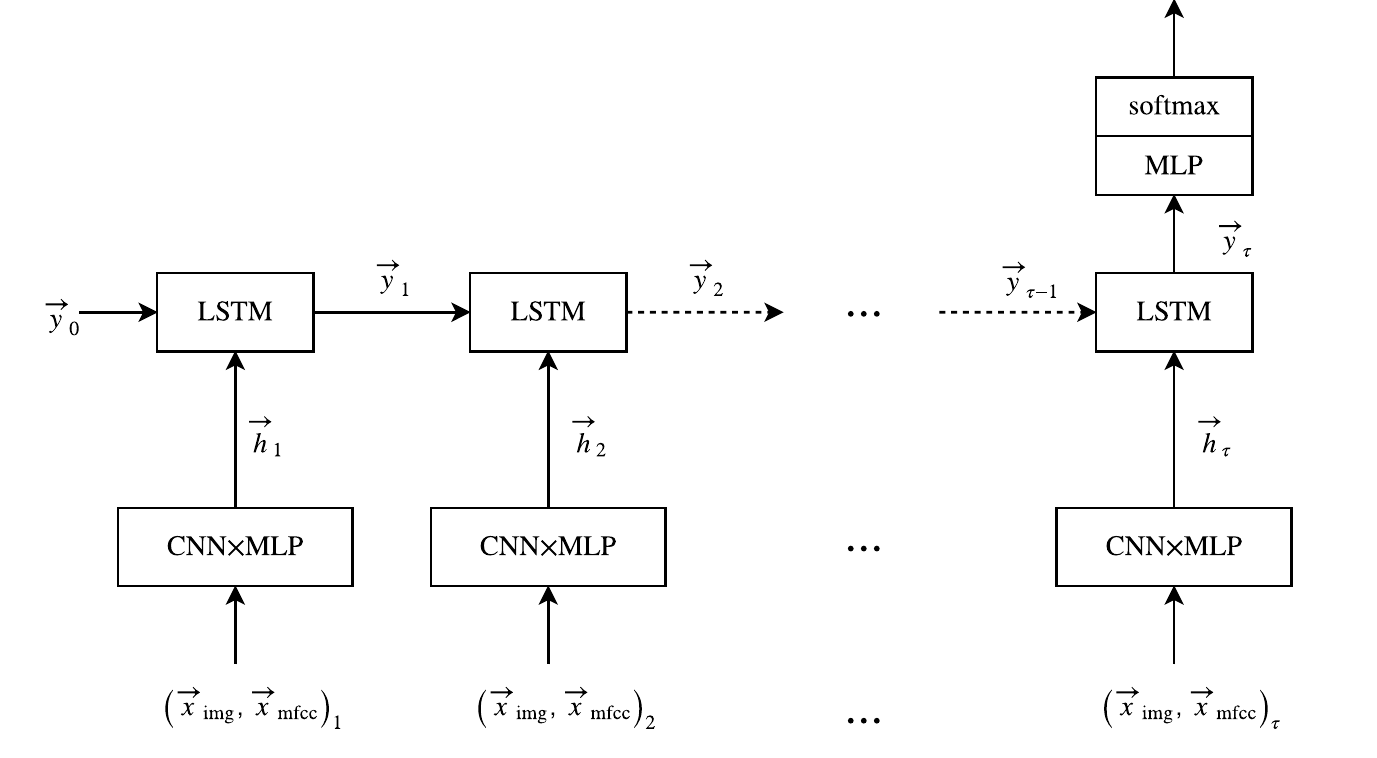}
\caption{\{CNN $\times$ MLP\}--LSTM macro-scale: sequential processing across time steps. The ``CNN $\times$ MLP'' rectangle represents the micro-scale per-frame extractor, shown in Figure~\ref{fig:f_ex}. The two input modalities are denoted by $\vec{x}_{\text{img}}$ and $\vec{x}_{\text{mfcc}}$, while $\vec{y}_t$ is the output of the LSTM layer at time $t$.}
\label{fig:lstm_seq}
\end{center}
\vskip -0.1in
\end{figure}

\begin{figure}[h]
\begin{center}
\includegraphics[clip, trim=1cm 8cm 5cm 1cm, width=0.35\textwidth]{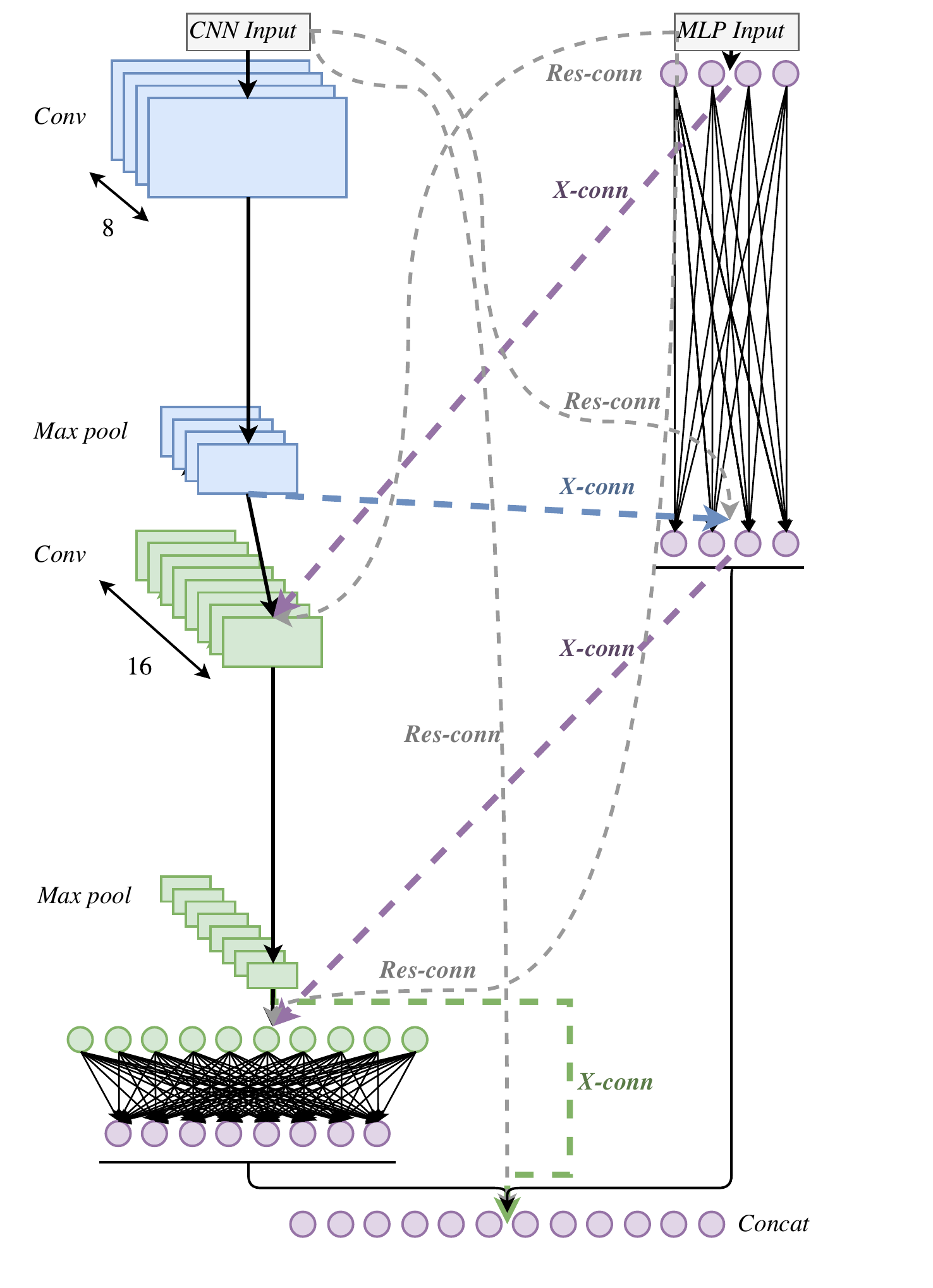}
\caption{CNN $\times$ MLP micro-scale: per-frame feature extractor with cross- and residual connections.}
\label{fig:f_ex}
\end{center}
\vskip -0.1in
\end{figure}

Shown in Figure~\ref{fig:f_ex}, the feature extractor for a single frame is weight-shared across all frames, which allows it to process input sequences of arbitrary lengths. After one set of features is extracted from the two modalities for each frame, it gets passed to an LSTM layer as an element $\vec{h}_i$ from the whole sequence, as illustrated in Figure~\ref{fig:lstm_seq}. This layer then produces a set of 64 features for the entire example which is finally classified by the softmax layer. Additionally, all cross- and residual connections are designed in the same manner as for the CNN $\times$ MLP architecture, but differ in the sense that they only operate within the space of the single-frame feature extractor.

While similar to the CNN $\times$ MLP model, this architecture differs in that the second convolutional layer from each \{conv$\times 2$, max-pool\} block has been removed and the number of kernels from the remaining layers has been halved. The underlying motivation for this choice arises from the features no longer being extracted from an averaged block corresponding to an entire video, but rather from an individual frame, thereby heavily sparsifying the available information for a single input.

\subsection{Model Architectures}\label{march}

Tables~\ref{table:avlettersarch} and~\ref{table:digitsarch} summarise the two models in terms of the number of parameters and cross-connections. For brevity, we have excluded the descriptions of residual connections, as they can be inferred from the shape of their target.

All convolutional and fully-connected layers in the architectures have ReLU activations. In terms of regularisation techniques for reducing overfitting in the CNN $\times$ MLP model, batch normalisation~\cite{ioffe2015batch} is applied after the input layer, each pair of convolutional layers, the first fully-connected layer in the MLP stream and the merge layer. We also applied dropout~\cite{srivastava2014dropout} with $p = 0.25$ after every max-pooling layer and with $p = 0.5$ after the first fully-connected layer in the MLP stream, the merge layer and the final fully-connected layer. We chose a larger value for $p$ in the latter case, due to the increased likelihood of overfitting in fully-connected layers, where the number of parameters is much larger than for convolutional layers. The \{CNN $\times$ MLP\}--LSTM model only employs batch normalisation after the input layer and merge layer, followed in the latter case by dropout with $p = 0.5$.

\begin{table}[h]
\caption{Architecture for the CNN $\times$ MLP baseline and model with cross-connections (whose parameters are described in {\bf bold}).}
\label{table:avlettersarch}
\begin{center}
\begin{small}
\begin{sc}
\setlength{\extrarowheight}{1.5pt}
\vskip 0.15in
\resizebox{\linewidth}{!}{%
\begin{tabular}{ c c c c } 
 \toprule
 Output size & CNN stream & MLP stream \\ 
 \midrule
 ([$80 \times 60$, 16], 128) & $[3 \times 3, 16]$ Conv $\times$ $2$ & Fully-connected 128-D \\
 ([$40 \times 30$, 16], 128) & $2 \times 2$ Max-Pool, stride $2$ &  \\
 \multirow{2}{*}{\bf([$\bf 40 \times 30, 32$], 192)} & \bf [$\bf 1 \times 1, 16$] Conv & \bf Fully-connected 759-D \\
 								  & \bf Fully-connected 64-D  $\bf \sersquigarrow$ & \bf $\bf \swlsquigarrow$ [$\bf 8 \times 8, 16$] Deconv \\
 
 ([$40 \times 30$, 32], 128) & $[3 \times 3, 32]$ Conv $\times$ $2$ & Fully-connected 128-D \\
 ([$20 \times 15$, 32], 128) & $2 \times 2$ Max-Pool, stride $2$ &  \\
 \multirow{2}{*}{\bf ([$\bf 20 \times 15, 64$], 256)} & \bf [$\bf 1 \times 1, 32$] Conv & \bf Fully-connected 204-D \\
 								  & \bf Fully-connected 128-D $\bf \sersquigarrow$ & \bf $\bf \swlsquigarrow$ [$\bf 4 \times 4, 32$] Deconv \\
 
 (256, 128) & Fully-connected 256-D & \\
 512 & \multicolumn{2}{c}{Fully-connected 512-D}  \\
  & \multicolumn{2}{c}{26-way softmax}  \\
 \bottomrule
\end{tabular}}
\end{sc}
\end{small}
\end{center}
\vskip -0.1in
\end{table}

\begin{table}[h]
\caption{Architecture for the \{CNN $\times$ MLP\}--LSTM baseline and model with cross-connections (whose parameters are described in {\bf bold}).}
\label{table:digitsarch}
\begin{center}
\begin{small}
\begin{sc}
\setlength{\extrarowheight}{1.5pt}
\vskip 0.15in
\resizebox{\linewidth}{!}{%
\begin{tabular}{ c c c c } 
 \toprule
 Output size & CNN stream & MLP stream \\ 
 \midrule
 ([$80 \times 60$, 8], 32) & $[3 \times 3, 8]$ Conv & Fully-connected 32-D \\
 ([$40 \times 30$, 8], 32) & $2 \times 2$ Max-Pool, stride $2$ &  \\
 \multirow{2}{*}{\bf([$\bf 40 \times 30, 8$], 64)} & \bf [$\bf 1 \times 1, 8$] Conv & \bf Fully-connected 375-D \\
 								  & \bf Fully-connected 32-D  $\bf \sersquigarrow$ & \bf $\bf \swlsquigarrow$ [$\bf 16 \times 16, 8$] Deconv \\
 
 ([$40 \times 30$, 16], 32) & $[3 \times 3, 16]$ Conv & Fully-connected 32-D \\
 ([$20 \times 15$, 16], 32) & $2 \times 2$ Max-Pool, stride $2$ &  \\
 \multirow{2}{*}{\bf ([$\bf 20 \times 15, 64$], 96)} & \bf [$\bf 1 \times 1, 16$] Conv & \bf Fully-connected 104-D \\
 								  & \bf Fully-connected 64-D $\bf \sersquigarrow$ & \bf $\bf \swlsquigarrow$ [$\bf 8 \times 8, 16$] Deconv \\
 
 (64, 32) & Fully-connected 64-D & \\
 64 & \multicolumn{2}{c}{LSTM}  \\
  & \multicolumn{2}{c}{26-way softmax}  \\
 \bottomrule
\end{tabular}}
\end{sc}
\end{small}
\end{center}
\vskip -0.1in
\end{table}

\vfill

We have observed that a XFlow model is underregularised, compared to its baseline under the same regularisation parameters. Due to the increase in input size to the following layer, all merging points are passed through a dropout layer with $p = 0.5$. Additionally, when transmitting data across streams, we have taken steps to ensure integrity of the information. This is where the ReLU activation experiences a shortcoming---approximately half of its outputs are zero upon Xavier~\cite{glorot2010understanding} initialisation. To enable the network to benefit from all transmitted data, we have applied the more general PReLU activation function~\cite{he2015delving}, which allows for data to ``leak'' in the negative input space.

\section{Experiments}

\subsection{Experimental Setup}

\subsubsection{Datasets}

We evaluated the performance of our models on the \emph{AVletters}~\cite{avlettersdataset} and \emph{CUAVE}~\cite{Patterson02cuave:a} benchmark tasks and on our own novel dataset \emph{Digits}. Additionally, in order to illustrate the improvements that our models can achieve by explicitly enforcing feature exchange, we also compare them to \emph{weakly-shared deep transfer networks}~\cite{shu2015weakly}, which allow interaction of parameters and representations across streams via custom loss functions.

Spanning 26 classes representing the letters A-Z, \emph{AVletters} contains 780 examples of 10 people saying each of the letters 3 times. We used two modalities from this dataset to construct the training and test examples: (1) audio data in the form of 1D mel-frequency cepstral coefficients (MFCCs) and (2) image data representing the video frames. The data was split into $k = 10$ folds to assess the performance of each classifier. Each of the folds corresponds to a different person in the dataset and can be seen as an extension to the usual leave-one-out cross-validation (LOOCV) approach, where each fold corresponds to one example. This allowed us to examine how well the models behave in a realistic audiovisual recognition setting---if we train a classifier with data collected from a group of people, we expect the model to be able to correctly identify the same information when being exposed to a new person.

The \emph{CUAVE} dataset has 10 classes (digits 0--9) and contains videos of 36 people saying each of the digits 5 times. In this case, we used the pre-processing described by Ngiam et al.~\cite{ngiam2011multimodal} and split the data into $k = 9$ folds, each one containing 4 people saying each digit 5 times.

Additionally, we have also curated a new dataset \emph{Digits}, suitable for audiovisual machine learning tasks. We have recorded 15 people saying the digits 0--9 in five different ways: in a \textit{low}, \textit{normal} and \textit{loud} voice, \textit{slowly} and \textit{quickly}. As a result, the \emph{Digits} dataset contains 750 video clips from which we then extracted two different modalities:
\begin{enumerate}
\item image data corresponding to the 2D video frames;
\item audio data either represented by MFCCs or by 2D spectrograms.
\end{enumerate}

\begin{figure}[!htb]
\minipage{0.15\textwidth}
  \includegraphics[width=\linewidth]{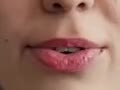}
  \label{fig:frame1}
\endminipage\hfill
\minipage{0.15\textwidth}
  \includegraphics[width=\linewidth]{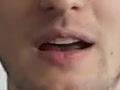}
  \label{fig:frame2}
\endminipage\hfill
\minipage{0.15\textwidth}%
  \includegraphics[width=\linewidth]{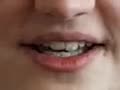}
  \label{fig:frame3}
\endminipage
\vskip -0.1in
\caption{A few examples of video frames from the \textit{Digits} dataset.}
\end{figure}

These three datasets present different characteristics: number of examples, classes and people present in the videos. 

Our dataset addresses the limitations of \emph{AVletters}---the lack of clear alignment between video and audio, which did not cause problems for prior research, but made it difficult to employ the kind of cross-modal feature passing described in this paper. This finding is reflected in all the results we report in subsequent tables, motivating the need for a novel benchmark.

\subsubsection{Optimisation}

Both XFlow architectures were trained using the Adam SGD optimiser for 300 epochs, with hyperparameters as described by Kingma and Ba~\cite{kingma2014adam} and a batch size of 128 for the CNN $\times$ MLP and 32 for the \{CNN $\times$ MLP\}--LSTM.

\subsubsection{Deep transfer network evaluation setup}

For evaluating the performance of weakly-shared deep transfer networks (DTNs) on this task, we trained a DTN with the following structure: $N_{\text{audio}} \rightarrow 512 \rightarrow 256 \rightarrow 256 \rightarrow 256 \rightarrow C$ and $N_{\text{image}} \rightarrow 2048 \rightarrow 256 \rightarrow 256 \rightarrow 256 \rightarrow C$, where $C =$ 10 (\emph{Digits}) or 26 (\emph{AVletters}, \emph{CUAVE}), $N_{\text{audio}} =$ 286 (\emph{AVletters}) or 156 (\emph{CUAVE} and \emph{Digits}) and $N_{\text{image}} =$ 13200 (\emph{AVletters}) or 16200 (\emph{CUAVE} and \emph{Digits}). In order to have a fixed-size input length, the DTN was given the same (flattened) data as the CNN $\times$ MLP. The batch size was 128 for all stages of evaluation.

The first two layers of each stream were pre-trained separately as single-layer autoencoders, until convergence. The entire DTN was trained using the $J_2$ loss over the entire dataset, since all examples can be considered co-occurrence pairs in our tasks. The values we found to work best for the other hyperparameters were: $\eta = 1\times10^{-6}$, $\gamma = 1\times10^{-2}$, $\lambda = 0$. Notably, other values for $\lambda$ did not seem to help the training of the DTN and a large ratio was required for $\eta$ and $\gamma$ to accommodate the relative magnitudes of the $J_2$ and $\Omega$ losses.

\subsubsection{Ablation study}

To gain a deeper understanding of the performance gains revealed by the cross-validation procedure, we investigate the improvements that our cross-modal connections bring individually to the XFlow architectures. For each of the scenarios mentioned in the experimental setup, we evaluate both CNN $\times$ MLP and \{CNN $\times$ MLP\}--LSTM models that only leverage cross-modal residual connections (denoted by \emph{No x-conns}) or cross-connections (\emph{No res-conns}).

\begin{figure}[h]
\begin{center}
\includegraphics[clip, trim=0cm 0cm 0cm 1cm, width=\linewidth]{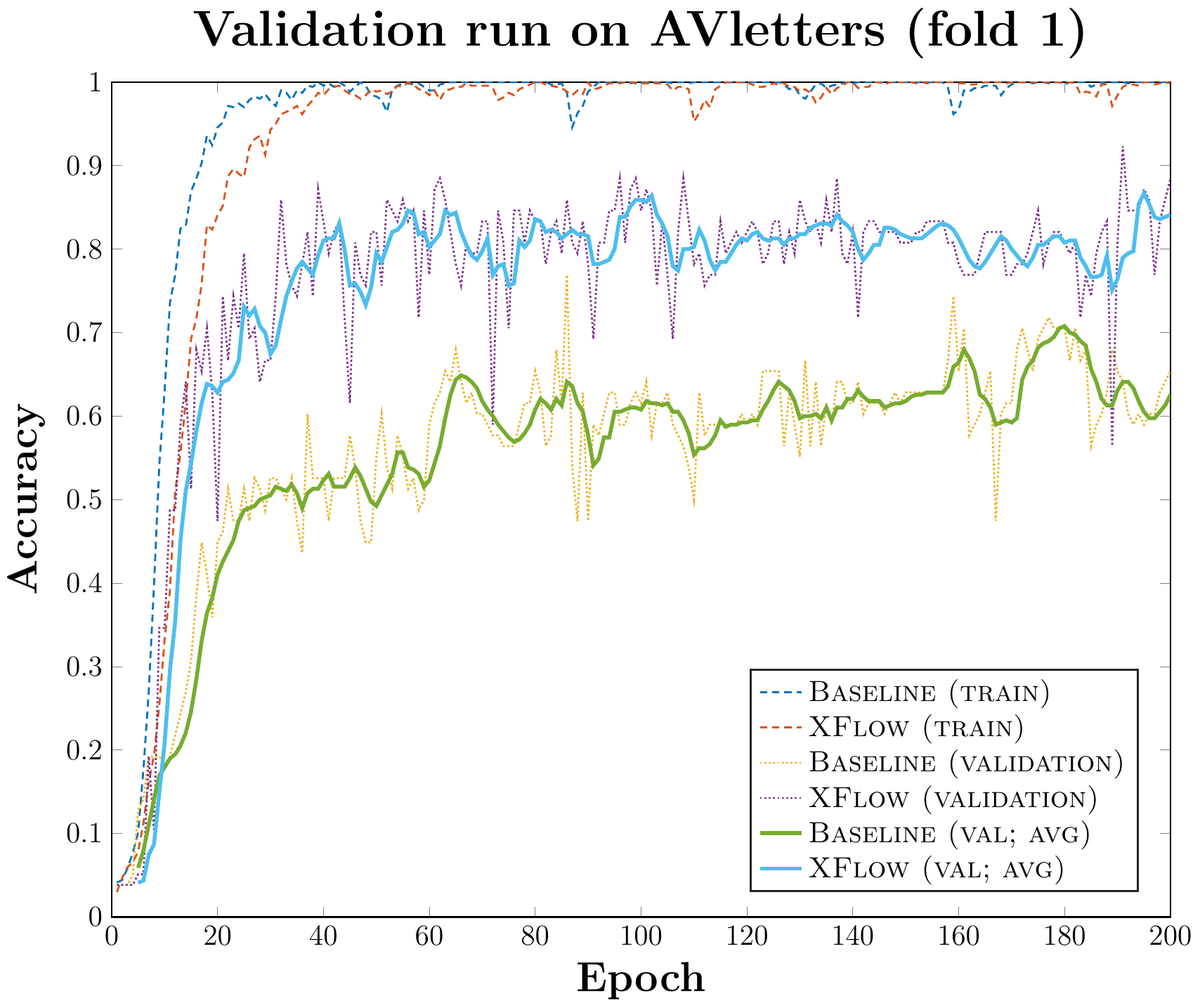}
\caption{Plot of the accuracy of the \{CNN $\times$ MLP\}--LSTM model on the first \emph{AVletters} cross-validation fold. We have also used a sliding averaging window of 5 epochs on the accuracy values, to emphasise the model capabilities during training.}
\label{fig:accuracy}
\end{center}
\vskip -0.1in
\end{figure}

\begin{figure}[h]
\begin{center}
\includegraphics[clip, trim=0cm 0cm 0cm 1cm, width=\linewidth]{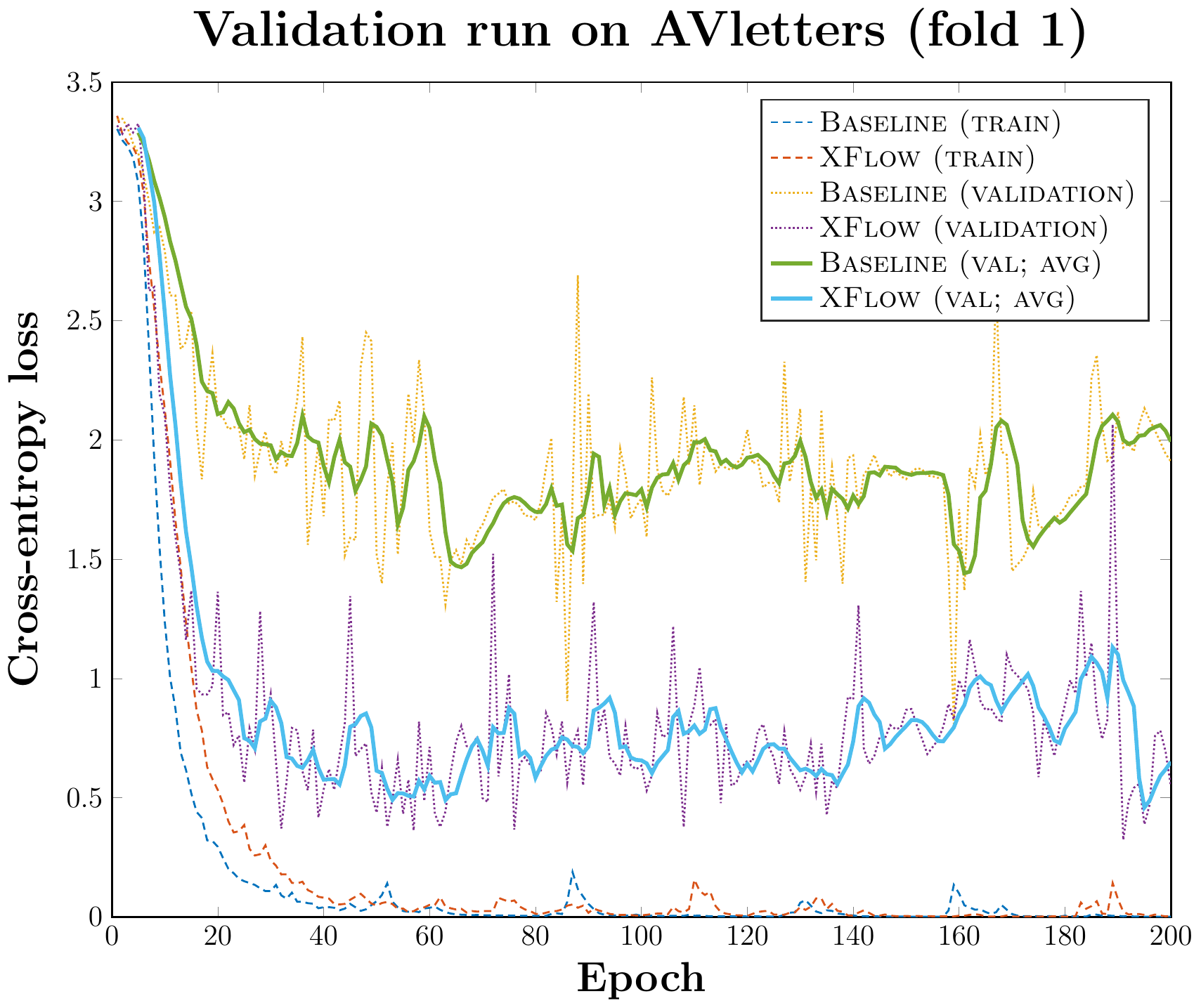}
\caption{Plot of the cross-entropy loss of the \{CNN $\times$ MLP\}--LSTM model on the first \emph{AVletters} cross-validation fold.}
\label{fig:xenloss}
\end{center}
\vskip -0.1in
\end{figure}

\subsection{Results}

\subsubsection{Comparison against baselines}

During evaluation, each architecture was tested using $k$-fold cross-validation for the \emph{AVletters} (10-fold), \emph{CUAVE} (9-fold) and \emph{Digits} (15-fold) datasets. However, some of the architectures we developed contain a large number of parameters (underlined in Table~\ref{table:params}~\footnote{Depending on the dataset, the cross-connected models differ in the number of parameters. This is due to the cross-connections encompassing operations such as transposed convolution, which requires its output data to be in a specific shape for concatenation with the other stream.}).

\begin{table}[htbp]
\caption{Number of trainable parameters in each considered model, for the \emph{AVletters} and \emph{Digits/CUAVE} datasets.}
\label{table:params}
\vskip 0.1in
\begin{center}
\begin{small}
\begin{sc}
\resizebox{\linewidth}{!}{%
\begin{tabular}{ c c c } 
 \toprule
 & AVletters & Digits/CUAVE \\ 
 \midrule
 Baseline \\
 \midrule
 CNN $\times$ MLP & \underline{2,740,512} & \underline{5,664,054} \\ 
 \{CNN $\times$ MLP\}--LSTM & 353,650 & 721,170 \\ 
 \midrule
 XFlow \\
 \midrule
 CNN $\times$ MLP & \underline{8,852,962} & \underline{17,764,002} \\ 
 \{CNN $\times$ MLP\}--LSTM & 1,387,488 & 2,967,388 \\ 
 \bottomrule
\end{tabular}}
\end{sc}
\end{small}
\end{center}
\vskip -0.1in
\end{table}

In such circumstances, initialisation heavily influences the representation learned during training in low-data scenarios (\emph{AVletters} and \emph{Digits}, which had approximately half as much training data as \emph{CUAVE}). Consequently, the accuracies obtained might not always reflect the best performance that the classifier has the potential to obtain on a particular validation fold. Because of this, we trained each model underlined in Table~\ref{table:params} five times per validation fold, for the two datasets, and recorded the maximum result. The final accuracy for an architecture was then computed as the average over all folds. For statistical significance testing, we used the paired $t$-test across fold results, with a significance threshold of $p \leq 0.05$.

\begin{table*}[h!]
\caption{Classification accuracy on \emph{AVletters}, \emph{Digits} and \emph{CUAVE} for the considered architectures, after running the $k$-fold cross-validation procedure. The $p$-values corresponding to statistically significant results (i.e. the XFlow model performs better than the baseline) are underlined---we used the paired $t$-test across results from all folds, with a significance threshold of $p \leq 0.05$.}
\label{table:results}
\vskip 0.1in
\begin{center}
\begin{small}
\begin{sc}
\resizebox{\linewidth}{!}{%
\begin{tabular}{lccccccccccr} 
 \toprule
 &\multicolumn{3}{c}{AVletters} &\multicolumn{3}{c}{Digits} &\multicolumn{3}{c}{CUAVE} \\ 
 & Baseline & XFlow & $p$-value & Baseline & XFlow & $p$-value & Baseline & XFlow & $p$-value \\ 
 \midrule
 DTN~\cite{shu2015weakly} &35.0\% & -- & -- & 72.9\% & -- & -- &87.9\% & -- & -- \\
 CNN $\times$ MLP & 73.1\% & \bf{74.0\%} & 0.65 & 78.3\% & \bf{87.7\%} & \underline{2 $\times 10^{-3}$} & 90.3\% & \bf{93.5\%} & \underline{0.05}  \\ 
 \{CNN $\times$ MLP\}--LSTM & 78.1\% & \bf{89.6\%} & \underline{0.02} & 88.7\% & \bf{96.8\%} & \underline{1.2 $\times 10^{-3}$} & 96.9\% & \bf{99.3\%} & \underline{0.01} \\
 \bottomrule
\end{tabular}}
\end{sc}
\end{small}
\end{center}
\vskip -0.1in
\end{table*}

The results for all classifiers are shown in Table~\ref{table:results}. The \{CNN $\times$ MLP\}--LSTM performed significantly better on the \emph{AVletters} dataset, with an improvement of 11.5\% over its baseline. The resulting $p$-value was 0.02, which corresponds to a 98\% confidence interval. However, the CNN $\times$ MLP only achieved a 0.9\% edge over the model without cross-connections, its $p$-value not showing statistical significance. Since the same model performed 9.4\% better than its baseline on the \emph{Digits} dataset, the explanation is clearly linked to the quality of the \emph{AVletters} dataset. The apparent discrepancy in visual and audio data pre-processing of this dataset required us to average MFCC sets and video frames over time windows of different lengths, which is likely to have resulted in misalignment of the visual and audio information. This, along with the averaging process that results in loss of information, led to a weaker performance of the CNN $\times$ MLP on \emph{AVletters}.

A highly significant performance gain of cross-connected models was achieved on the \emph{Digits} dataset, the highest $p$-value being $2 \times 10^{-3}$. Adding cross-connections in the CNN $\times$ MLP boosted performance the most, with an improvement of 11.5\%. Finally, the most impressive result is achieved by the \{CNN $\times$ MLP\}--LSTM architecture with an accuracy of 96.8\%. It is also worth noting that this model has correctly classified all examples from one fold and that there are remarkable benefits to temporal sequence modelling, as the corresponding baseline performed better than the CNN $\times$ MLP on this dataset, obtaining an overall accuracy of 88.7\%, all of this achieved by using more than 5 times fewer parameters than any of them. Similar outperformance has been observed on the \emph{CUAVE} dataset, with $p$-values indicating that the XFlow models improve on their respective baselines with statistical significance.

Finally, we note that the weakly-shared DTN fails to match the performance of the CNN $\times$ MLP, for all three datasets. The largest difference in performance can be noticed on \emph{AVletters}, which once again suggests a considerably lower data quality, in terms of misalignment and over-processing. This stops the correspondence across modalities to be more usefully exploited.

The plots in Figures~\ref{fig:accuracy} and~\ref{fig:xenloss} show how the validation accuracy and cross-entropy loss, respectively, evolve as a function of the training epoch. A significant improvement over the baseline (same model, without the cross-modal connections) can be seen in both plots.

The results outlined above successfully demonstrate that statistically significant benefits can be obtained by exploiting the nontrivial cross-modality during the feature extraction stage. It is also expected that these findings should easily generalise to domains beyond audiovisual classification, given the general end-to-end structure and training of the networks involved.

\subsubsection{Ablation study}

\begin{table*}[t]
\caption{Classification accuracy on \emph{AVletters}, \emph{Digits} and \emph{CUAVE} for ablated XFlow models, after running the $k$-fold cross-validation procedure as described earlier in this section. For each of the benchmarks, we present the model performance when both types of connections are present in the architecture, then accuracies for each of the ablated models (no cross-connections, no residual cross-modal connections).}
\vspace{-1em}
\label{table:ablations}
\vskip 0.1in
\begin{center}
\begin{small}
\begin{sc}
\resizebox{\linewidth}{!}{%
\begin{tabular}{lccccccccccr} 
 \toprule
 &\multicolumn{3}{c}{AVletters} &\multicolumn{3}{c}{Digits} &\multicolumn{3}{c}{CUAVE} \\ 
 & Both & No x-conns & No res-conns & Both & No x-conns & No res-conns & Both & No x-conns & No res-conns \\ 
 \midrule
 CNN $\times$ MLP & \bf 74.0\% & 73.8\% & 70.6\% & 86.7\% & \bf 87.7\% & 80.1\% & \bf 93.5\% & 92.0\% & 91.9\%  \\ 
 \{CNN $\times$ MLP\}--LSTM & 85.6\% & \bf 89.6\% & 86.2\% & 93.0\% & \bf 96.8\% & 94.1\% & 98.8\% & \bf 99.3\% & 97.7\% \\
 \bottomrule
\end{tabular}}
\end{sc}
\end{small}
\end{center}
\vskip -0.1in
\end{table*}

We report the results of these experiments in Table~\ref{table:ablations}. It is immediately obvious that cross-modal residual connections play an essential role in improving the performance of both models---removing them results in the biggest accuracy drop, while keeping them as the only means of cross-modal exchange often results in an even better accuracy than using both kinds of connections (on both models, in the case of \emph{Digits}, and on the recurrent model, for the other two benchmarks). This suggests that a network might find it easier to freely decide how to process the raw input data, rather than having to make use of intermediate features from the other stream.

Furthermore, the recurrent model appears to perform better when only leveraging one type of connection, regardless of whether it is an inner or a residual cross-modal connection, for \emph{AVletters} and \emph{Digits}. This could imply that representations encoded by different types of connections might be more difficult to integrate at a single merging point in the network, and that a relatively lightweight XFlow model can perform better in the context of an audiovisual task with aligned data.

By exploring various ways of exchanging information using the proposed cross-modal connections, we therefore find the best-performing model for each of the audiovisual tasks.

\subsubsection{State-of-the-art comparison}

In addition to the cross-validation method described previously, we have compared the performance of the \{CNN $\times$ MLP\}--LSTM models against \emph{CorrRNN}~\cite{YangRCMBL17}, the latest published state-of-the-art result on \emph{AVletters} and \emph{CUAVE} (to the best of our knowledge). Influenced by the aforementioned cross-validation results, we have decided to use only the recurrent model for comparison. For fairness, we have used the same train/test partition as detailed by Ngiam et al.~\cite{ngiam2011multimodal}, used by all previous published approaches. The results are summarised in Table~\ref{table:stateoftheart}.

\begin{table}[h!]
\caption{Comparative evaluation results of the \{CNN $\times$ MLP\}--LSTM models against the state-of-the-art approach on \emph{AVletters} and \emph{CUAVE}, using the holdout method described by Ngiam et al.~\cite{ngiam2011multimodal}.}
\vspace{-1em}
\label{table:stateoftheart}
\vskip 0.1in
\begin{center}
\begin{small}
\begin{sc}
\resizebox{\linewidth}{!}{%
\begin{tabular}{ c c c }
 \toprule 
 & AVletters & CUAVE \\ 
 \midrule
 CorrRNN~\cite{YangRCMBL17} & 83.4\% & 95.9\% \\ 
 \{CNN $\times$ MLP\}--LSTM (Baseline) & 91.5\% & 96.1\% \\ 
 \{CNN $\times$ MLP\}--LSTM (XFlow) & \bf{94.6\%} & \bf{96.9\%} \\ 
 \bottomrule
\end{tabular}}
\end{sc}
\end{small}
\end{center}
\end{table}

Evidently, the prior state-of-the-art approach is outperformed by both the baseline and XFlow models. The relative test error is improved by 67.5\% on \emph{AVletters} and by 24.4\% on \emph{CUAVE}.

\subsection{Interpretability of Cross-connections}

\subsubsection{2D$\rightsquigarrow$1D transformations}
Evaluation results showed that cross-connections significantly improve the performance of their baseline models. This means that they eventually lead to a better discrimination ability, so the outputs of cross-connections for particular training examples might exhibit some form of clustering, according to the classes the examples belong to.

Using a pre-trained CNN $\times$ MLP model, we chose one example from each person and class in the \emph{Digits} dataset and provided each of the examples as input to the architecture. We then looked at the outputs of the 2D$\rightsquigarrow$1D cross-connections within the CNN $\times$ MLP---since this model processes an input corresponding to a whole video, the 1D data comes in the form of a single output vector per example and can be more easily interpreted. To illustrate the potential clustering, we have used the $t$-SNE \cite{maaten2008visualizing} algorithm for visualising high-dimensional data in two or three dimensions.

\begin{figure}[h]
\begin{center}
\includegraphics[width=0.47\textwidth]{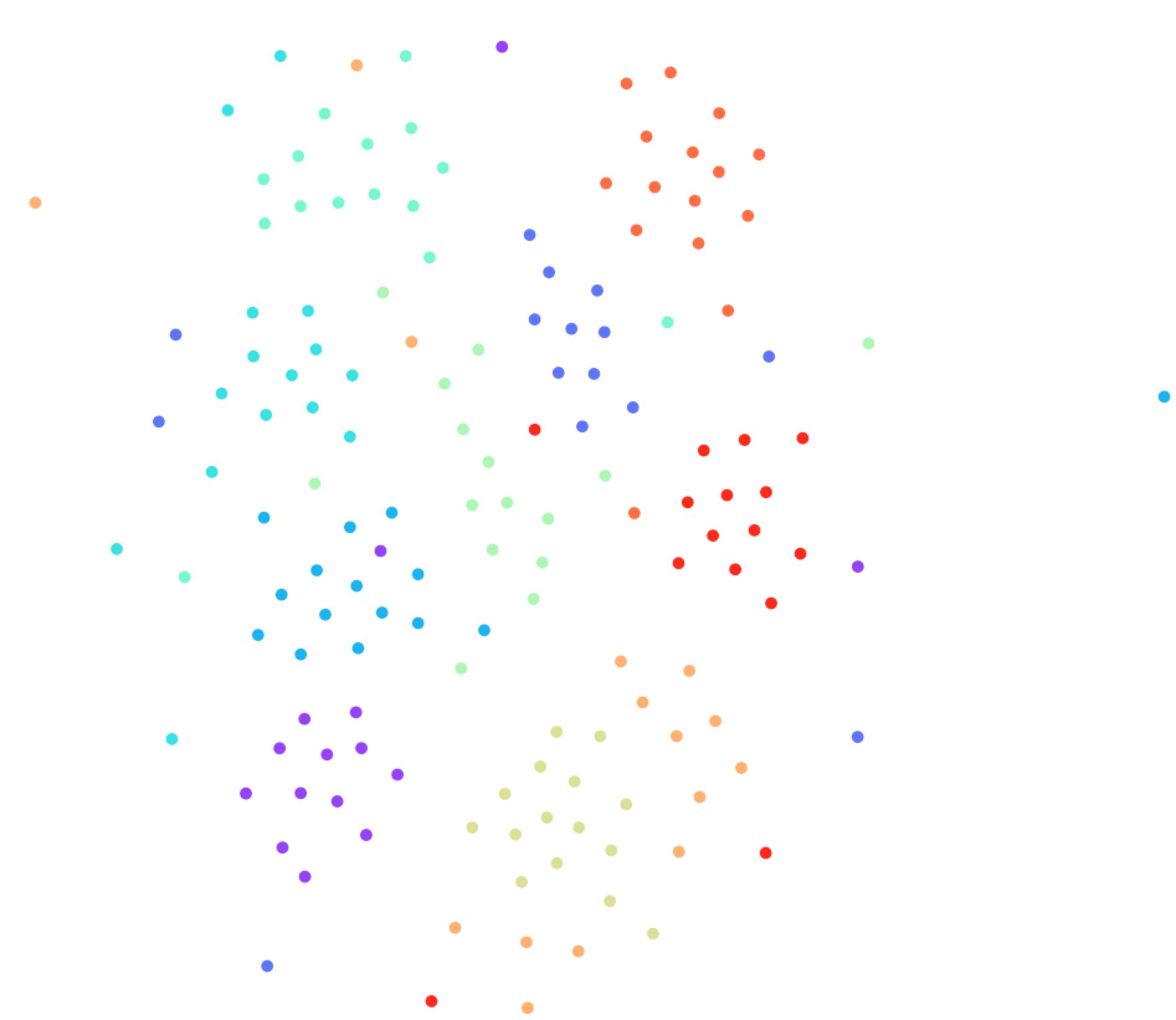}
\caption{Best viewed in colour. Two-dimensional $t$-SNE plot of the outputs of the second 2D$\rightsquigarrow$1D cross-connection within the CNN $\times$ MLP model. Each colour corresponds to a different class from the \emph{Digits} dataset.}
\label{fig:tsne}
\end{center}
\vskip -0.1in
\end{figure}

With the exception of a few outliers for each class, there is a discernible clustering (Figure~\ref{fig:tsne}). This shows that two intermediate representations resulting from input examples belonging to the same class are processed in a more similar manner by the second 2D$\rightsquigarrow$1D cross-connection. The same process was repeated for the first cross-connection within the same model, but no apparent clustering was identified. Since this means that the features were extracted from the 2D stream at an earlier point, they are likely to have still been in their primitive stages and therefore not discriminative enough across the 10 classes.

\subsubsection{1D$\rightsquigarrow$2D transformations}

A different kind of analysis was required for connections going in the opposite direction, transforming 1D data to 2D representations. Several cross-connections (both inter-stream and residuals) were analysed. The most revealing results were obtained from the \{CNN $\times$ MLP\}--LSTM architecture, which models temporal structure and can therefore produce a sequence of connection outputs.

We have visualised the transformations performed by the first residual connection that processes the MLP input and sends it to the CNN---for each frame, this directly turns a vector of 26 MFCCs into a 2D image. For the residual connections to be helpful, we expected them to be able to preserve the changes occurring across time in the audio signal, while producing images of well-defined structure for the 2D stream.

For an input sequence of MFCC sets and the corresponding outputs of the residual connection, we have quantified the relative changes between consecutive time steps. The $L^2$ norm (Euclidean distance) has been used to calculate the differences for both input vectors and resulting 2D matrices, by summing over the squared element-wise differences:

\begin{equation}
\begin{aligned}
{\textit{diff}_{\text{MFCC}}}_t &= \parallel \mathbf{x}_t - \mathbf{x}_{t - 1} \parallel^2,\\
{\textit{diff}_{\text{img}}}_t &= \parallel \text{res\_conn}_1(\mathbf{x}_t) - \text{res\_conn}_1(\mathbf{x}_{t-1}) \parallel^2.
\end{aligned}
\label{eq:diff}
\end{equation}

\begin{figure}[h]
\begin{center}
\setlength{\tabcolsep}{0.2em}
\begin{tabular}{c c}
\includegraphics[height = 0.11\textheight]{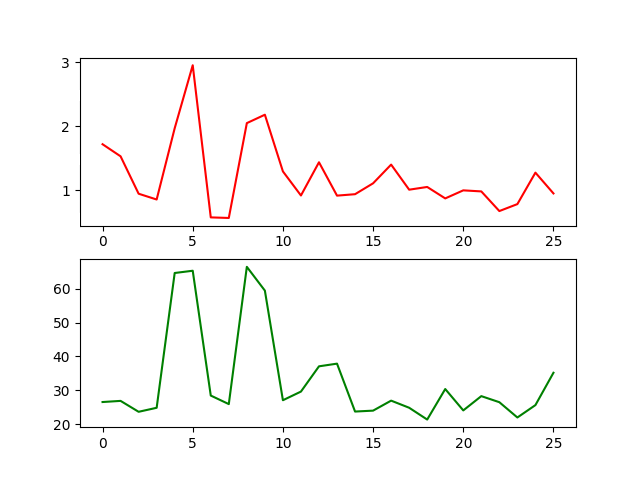} &
\includegraphics[height = 0.11\textheight]{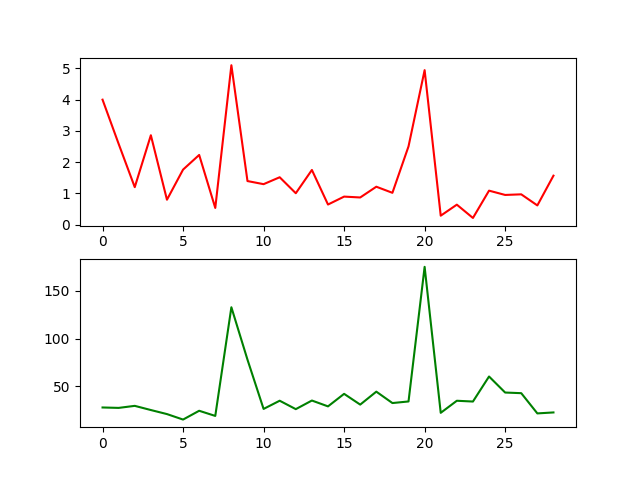} \\
\includegraphics[height = 0.11\textheight]{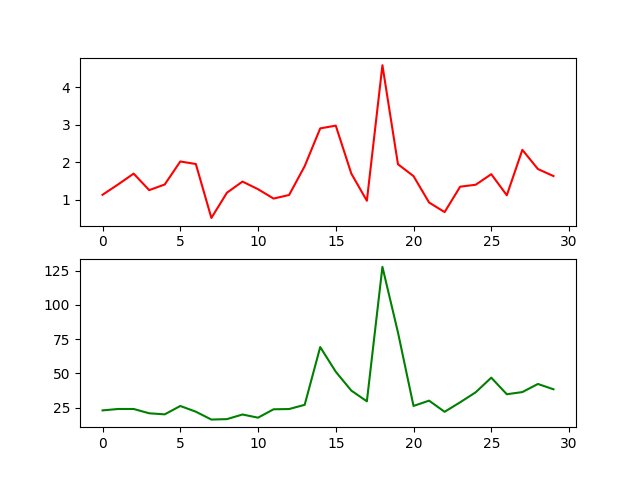} &
\includegraphics[height = 0.11\textheight]{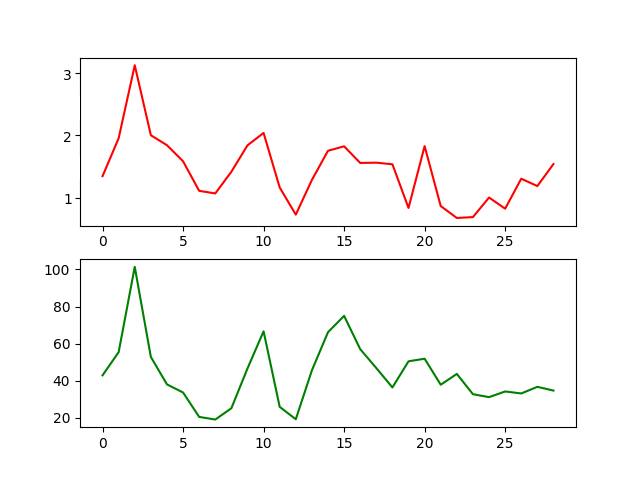} \\
\includegraphics[height = 0.11\textheight]{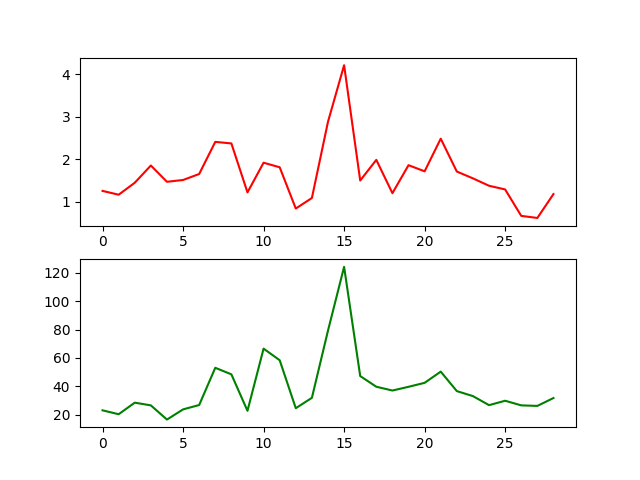} &
\includegraphics[height = 0.11\textheight]{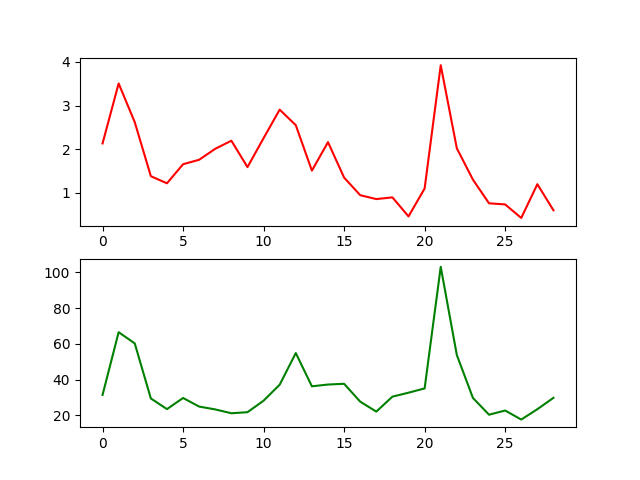} \\
\end{tabular}
\end{center}
\caption{Differences computed as described in equation~\ref{eq:diff}, for 6 of the kernels in the final (deconvolution) layer within the first \{CNN $\times$ MLP\}--LSTM 1D$\rightsquigarrow$2D residual connection and one \emph{Digits} example per kernel. The horizontal axis represents time, the vertical one represents the value of the $L^2$ norms. \emph{(Top:)} Differences between the residual connection outputs. \emph{(Bottom:)} Corresponding differences for the MLP 1D inputs.}
\label{fig:resdiff}
\end{figure}

The plots in Figure~\ref{fig:resdiff} illustrate that the first 1D-2D residual connection manages to preserve the changes that occur across time frames in the MLP input. This shows that these non-trivial residual cross-connections are performing a transformation capable of conveying useful information to the other stream.

\begin{figure}[h]
\begin{center}
\setlength{\tabcolsep}{0.2em}
\begin{tabular}{c c}
\includegraphics[height = 0.105\textheight]{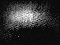} &
\includegraphics[height = 0.105\textheight]{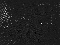} \\
\includegraphics[height = 0.105\textheight]{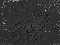} &
\includegraphics[height = 0.105\textheight]{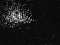} \\
\includegraphics[height = 0.105\textheight]{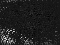} &
\includegraphics[height = 0.105\textheight]{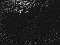} \\
\end{tabular}
\end{center}
\caption{Example outputs of the first \{CNN $\times$ MLP\}--LSTM 1D$\rightsquigarrow$2D residual connection, for the same 6 deconvolution layers as in Figure~\ref{fig:resdiff}. The feature maps illustrate how our model translates the structural information present in the 1D input when converting it to a 2D representation.}
\label{fig:resdiff2d}
\end{figure}

Furthermore, even though the 2D outputs shown in Figure~\ref{fig:resdiff2d} do not contain features that are interpretable by the human eye, it can be certainly observed that each feature map is still displaying a non-random pattern. This has very different characteristics from what would have resulted if no learning had taken place after the kernels' initialisation---a white noise image, essentially.

These visualisations prove that each modality can be helpfully converted to the other one within the architectures that we have designed. In the 2D-1D direction, the outputs visibly produce clustering according to the classes, whereas the 1D-2D transformation preserves the dynamics across the time steps for an entire video sequence, presenting these dynamics in a structured 2D manner. Both of these eventually lead to the observed improvement in classification accuracy.

While the above analyses represent a small step towards the general problem of neural network interpretability~\cite{lipton2016mythos}, the results observed are largely encouraging. In particular, residual cross-connections pave the way for a methodology that allows us to almost directly assess the way in which raw inputs of one kind relate to higher-level features of another kind, potentially allowing us to draw useful conclusions about cross-modal systems in general.

\vspace{-1em}

\section{Related Work}

Cross-modal algorithms have been constantly emerging over the past few years, often encountered in image and text settings. Tang et al.~\cite{tang2016generalized} used stacked denoising autoencoders to maximise correlations among modalities, at the same time extracting domain-invariant semantic features. Shu et al.~\cite{shu2015weakly} employ a deep transfer network with weakly parameter-shared layers to translate between the text to the image domain. They use an additional label transfer loss and regularisation to control the extent of parameter sharing. Building on the aforementioned work, Yang et al.~\cite{yang2015cross} learn semantic knowledge from web texts, which is then transferred to the image domain in the case of insufficient visual training data, weakly-sharing both parameters and representations of higher layers. Kang et al.~\cite{kang2015learning} also explore unpaired data settings, jointly learning the basis matrices of modalities from the training samples. A local group-based priori is used to learn the most discriminant groups for classification. This shrinks redundant and noisy groups, improving resistance to noise and intra-class variations. Aytar et al.~\cite{aytar2018cross} enable cross-modal convolutional networks to share representations in a modality-agnostic fashion. They encourage corresponding layers to have similar statistics across modalities via a multivariate Gaussian- or Gaussian mixture-parameterised regularisation term over network activations from PlacesCNN. Multimodal tasks are viewed from a generative perspective in some cases, such as the work of Gu et al.~\cite{gu2018look}, where image-to-text and text-to-image generative processes are incorporated in the cross-modal feature extraction. Both global abstract and local grounded features can thus be learned in a max-margin learning-to-rank framework.

Applications of multimodal learning are plentiful---for example, Salvador et al.~\cite{salvador2017learning} find a joint embedding of cooking recipes and images, regularizing the network via a high-level classification objective. This ensures that both recipe and image embeddings utilise the high-level discriminative weights in a similar way, adding a discrimination-based alignment degree to the process. Dou et al.~\cite{dou2018unsupervised} use unsupervised adversarial learning to tackle image segmentation in the medical domain. They transfer a convolutional neural network from the source (MRI) to the target (CT) images domain, using a domain adaptation module (DAM) that maps the target input data to the source feature space. Along with a domain critic module (CNN discriminator), the DAM is placed in a minimax two-player game used to train the framework. Albanie et al.~\cite{albanie2018emotion} develop a teacher network for facial emotion recognition and use it for training a student to learn speech embeddings in the absence of labels. This is achieved by transferring expression annotations from the visual domain (faces) to the speech domain (voices) through cross-modal distillation.

Various kinds of fusion are leveraged by Kiela et al.~\cite{kiela2015multi} for examining the grounding of semantic (textual) representations in raw audio data, in contexts such as zero-shot learning for mapping between linguistic and auditory data and unsupervised instrument clustering. Notably, even though ``early'', ``middle'' and ``late'' fusion methods are analysed, none of these involve an explicit information exchange between modalities. A more involved fusion approach is described by Gupta et al.~\cite{gupta2016cross}. They use the learned representations from a large labeled modality (RGB images) as the supervision signal for learning a new, paired unlabelled modality (depth and optical flow images), forcing corresponding layers in the model hierarchy to be similar via a Euclidean loss function. Even though this approach involves intra-layer feature exchange, it is limited to modalities that belong to the same (visual) data type.

Finally, Yang et al.~\cite{YangRCMBL17} tackle the audiovisual task presented here by developing a cross-modal approach that fuses input modalities with temporal structure. A maximum correlation loss term is used to facilitate cross-modal learning, while an attention model adjusts the contribution of modalities towards the joint representation. Once more, we note the lack of involved information exchange in earlier stages of the learning process, concluding the section by iterating that our model obtains superior performance through cross-modal feature exchanges that are concurrent with the feature extraction.

\section{Conclusion}

The purpose of this study was to design and validate a new method of performing cross-modal information transfer during the feature extraction process, within a classical multimodal deep learning architecture. Our primary contribution consists of newly developed cross- and residual connections that can transform 1D to 2D representations and vice versa, each of them being a sequence of neural network layers. These connections have been incorporated by two deep learning architectures for audiovisual data that could easily be generalised to other kinds of information. We emphasise and address the main challenge encountered in developing cross-modal connections---the fundamental incompatibility between the data types that are being exchanged.

Results show that our novel cross-modality enabled both architectures to favourably exploit the correlations between modalities, outperforming their corresponding baselines by up to 11.5\% on the AVletters, CUAVE and the novel Digits datasets. Comparative evaluation has also demonstrated that the \{CNN $\times$ MLP\}--LSTM models have achieved state-of-the-art performance on the AVletters and CUAVE datasets.

Another essential part of this research involved investigating the representations being learned by the new cross-modal connections. The 1D$\rightsquigarrow$2D residual connections in the \{CNN $\times$ MLP\}--LSTM preserved the intensity changes across the temporal axis when converting the 1D input into 2D information. Likewise, the second 2D$\rightsquigarrow$1D cross-connection in the CNN $\times$ MLP produces features that exhibit visible clustering.

Finally, we contribute to the machine learning research community by constructing Digits---a novel dataset that contains three different modalities (MFCCs, spectrograms and video frames). This data can be used in future multimodal research to evaluate new methods of performing cross-modality for various learning tasks.

\bibliographystyle{IEEEtran}
\bibliography{bare_jrnl}
%

%

\begin{IEEEbiography}[{\includegraphics[width=1in,height=1.25in,clip,keepaspectratio]{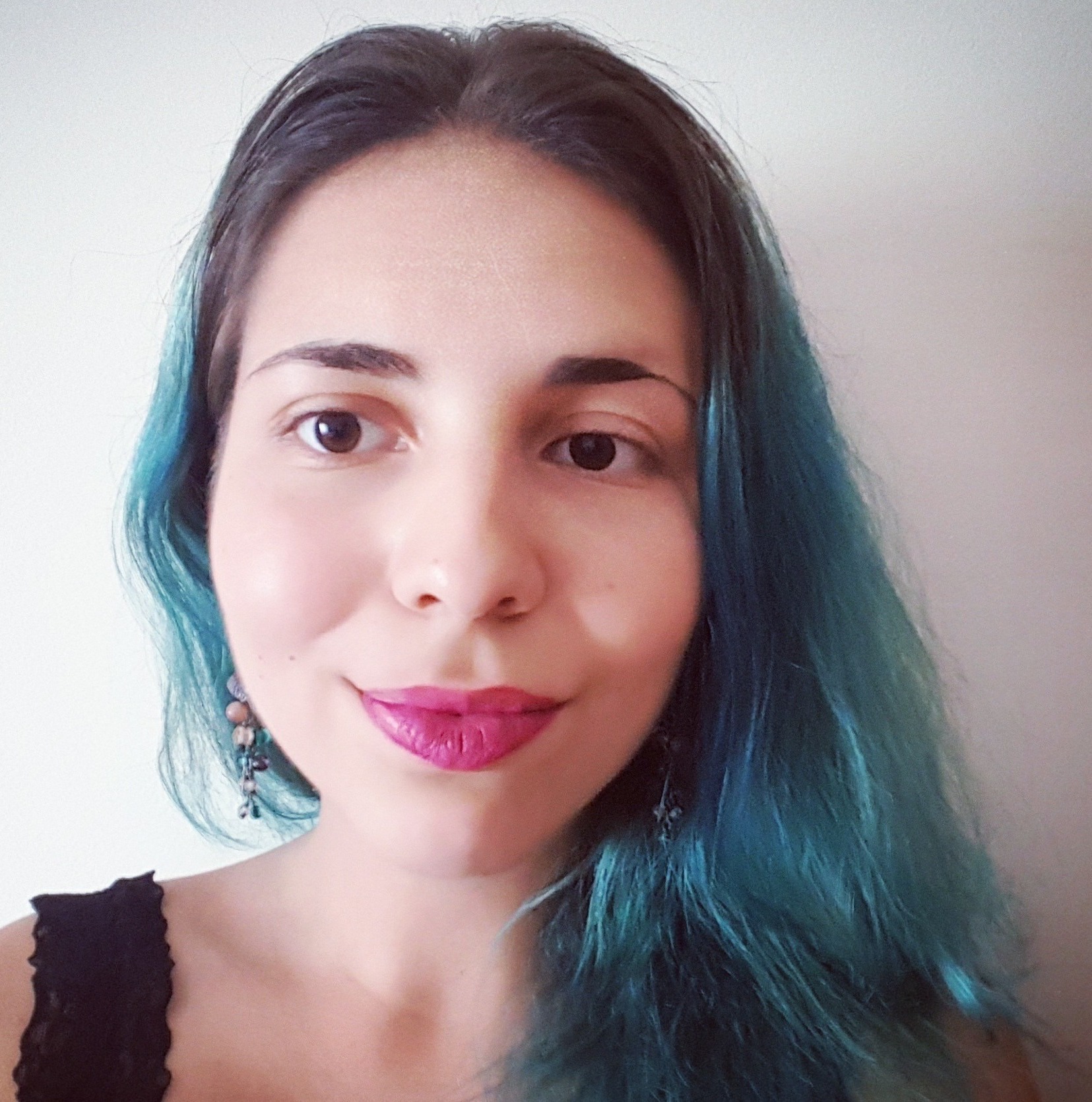}}]{C\u{a}t\u{a}lina~Cangea}
C\u{a}t\u{a}lina~Cangea is a second-year PhD student in Machine Learning at the Department of Computer Science and Technology of the University of Cambridge. She also holds a BA ('16) and an MPhil ('17) degree in Computer Science from the same department. Her research interests focus on integrating heterogeneous data with incompatible structures in multimedia and environmental scenarios, visual question answering and relational representation learning. She has presented her work at various machine learning venues (NeurIPS ML4H and R2L, WiML, IEEE ICDL-EPIROB Workshop on Computational Models for Crossmodal Learning) and events with broader scopes (ARM Research Summit). Currently, she is involved in a research collaboration with Aaron Courville at Mila Montr\'eal.
\end{IEEEbiography}

\vfill

\begin{IEEEbiography}[{\includegraphics[width=1in,height=1.25in,clip,keepaspectratio]{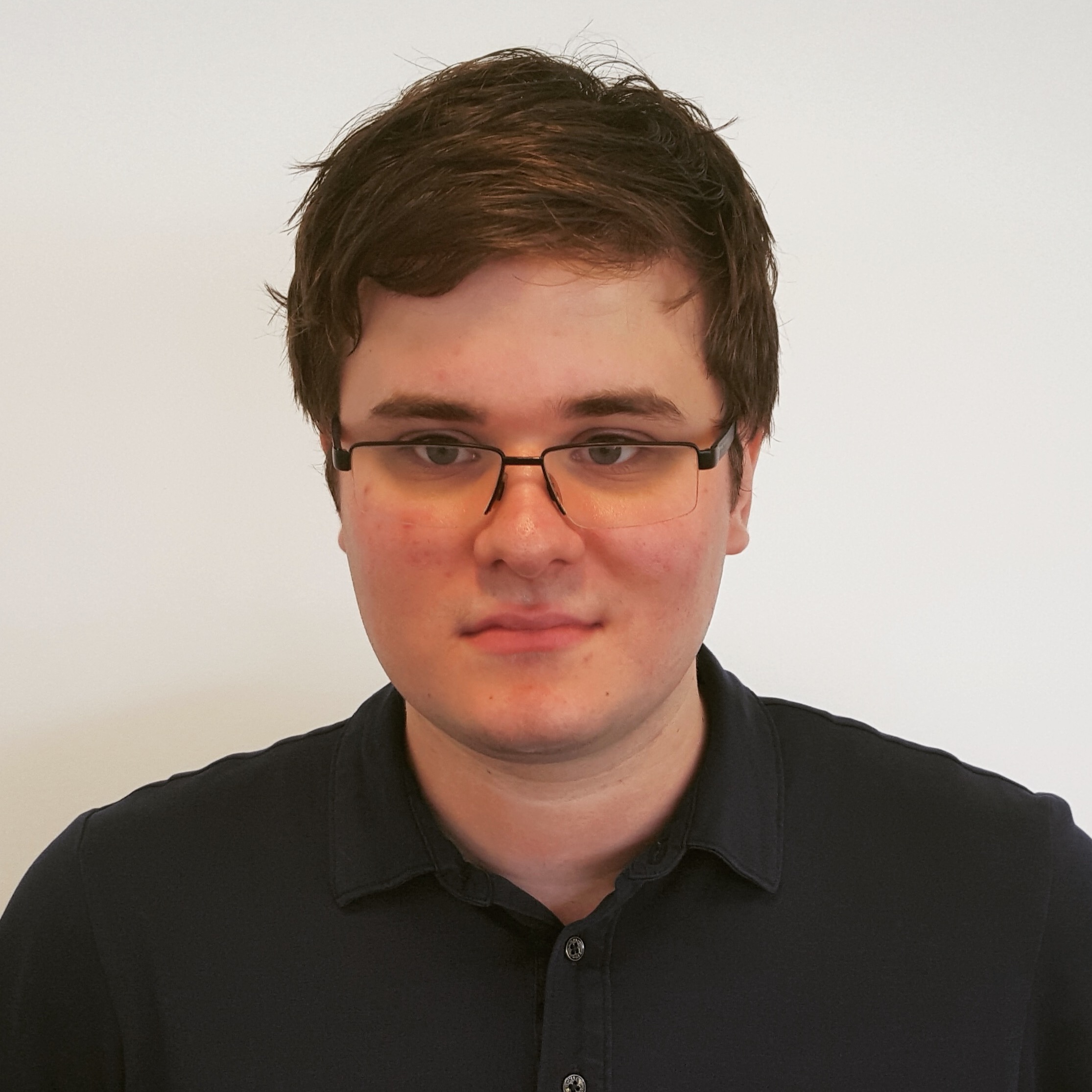}}]{Petar~{Veli{\v c}kovi{\'c}}}
Petar {Veli{\v c}kovi{\'c}} is currently a final-year PhD student in Machine Learning and Bioinformatics at the Department of Computer Science and Technology of the University of Cambridge. He also holds a BA degree in Computer Science from Cambridge, having completed the Computer Science Tripos in 2015. In addition, he has been involved in research placements at Nokia Bell Labs (working with Nicholas Lane) and the Montr\'{e}al Institute of Learning Algorithms (working with Adriana Romero and Yoshua Bengio). His current research interests broadly involve devising neural network architectures that operate on nontrivially structured data (such as graphs), and their applications in bioinformatics and medicine. He has published his work in these areas at both machine learning venues (ICLR, NeurIPS ML4H and R2L) and biomedical venues and journals (Bioinformatics, PervasiveHealth).
\end{IEEEbiography}

\vfill

\begin{IEEEbiography}[{\includegraphics[width=1in,height=1.25in,clip,keepaspectratio]{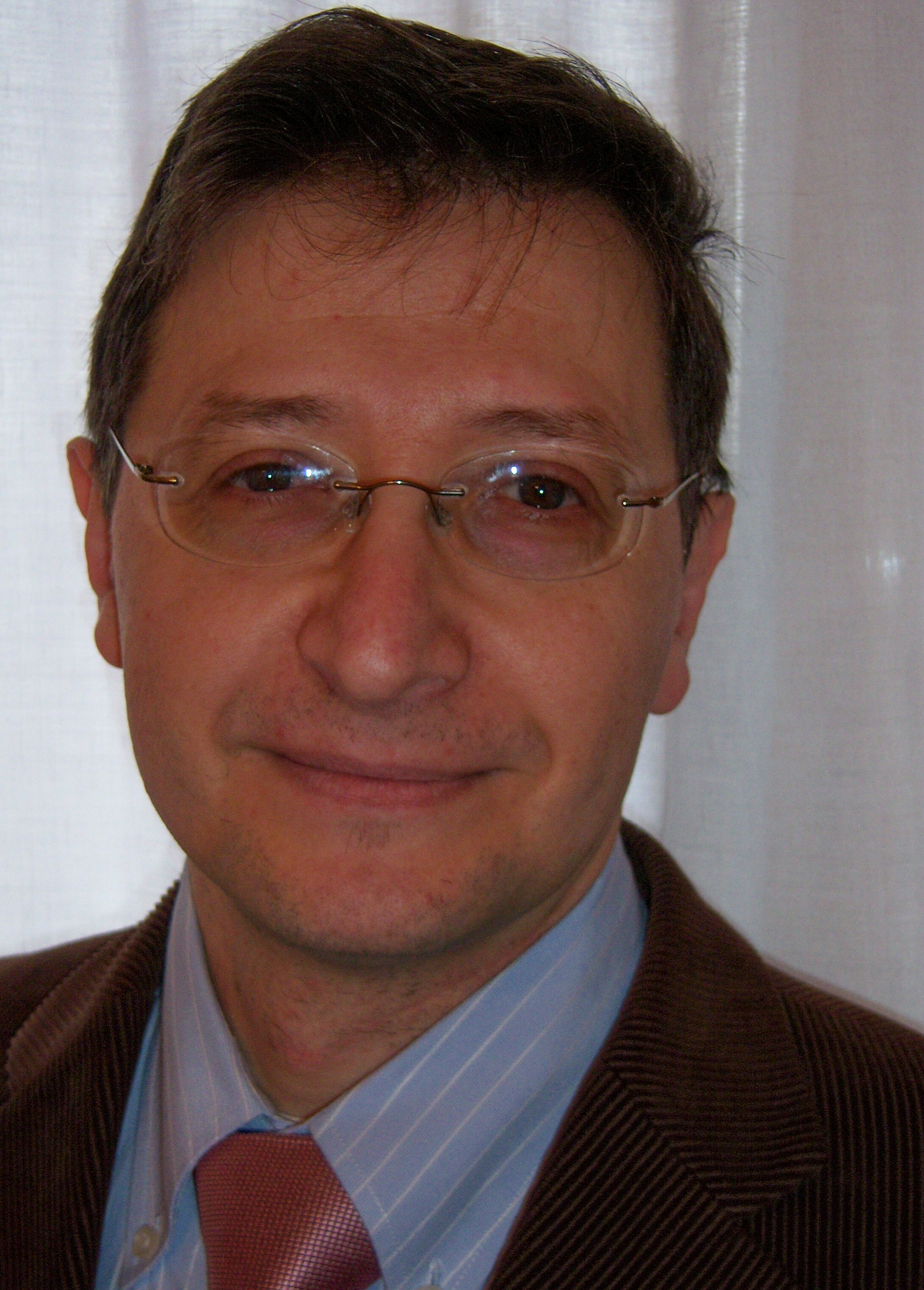}}]{Pietro~Li\`o}
Pietro Li\`o is a Professor in Computational Biology at the department of Computer Science and Technology of the University of Cambridge and he is a member of the Artificial Intelligence group of the Computer Laboratory. He has a MA from Cambridge, a PhD in Complex Systems and Non Linear Dynamics (School of Informatics, Department of Engineering of the University of Firenze, Italy) and a PhD in (Theoretical) Genetics (University of Pavia, Italy). His research areas include machine learning and data mining, data integration, computational models in health big data, and predictive models in personalised medicine.
\end{IEEEbiography}


\vfill


\end{document}